\documentclass[10pt,twocolumn,letterpaper]{article}

\usepackage{iccv}
\usepackage{times}
\usepackage{epsfig}
\usepackage{graphicx}
\usepackage{amsmath}
\usepackage{amssymb}
\usepackage{color,soul}
\usepackage{pdfpages}
\usepackage[noend]{algorithm2e}
\usepackage[dvipsnames]{xcolor}
\usepackage{color}
\usepackage{enumitem}
\setlist{nolistsep}
\usepackage{makecell}
\newsavebox\dotbox
\sbox{\dotbox}{\(\displaystyle\bigodot\)}

\usepackage{gensymb}
\usepackage{amsfonts}
\usepackage{multirow}

\usepackage[pagebackref=true,breaklinks=true,letterpaper=true,colorlinks,bookmarks=false]{hyperref}
\newcommand{\tabincell}[2]{\begin{tabular}{@{}#1@{}}#2\end{tabular}}

\usepackage[pagebackref=true,breaklinks=true,letterpaper=true,colorlinks,bookmarks=false]{hyperref}

\iccvfinalcopy 

\def\iccvPaperID{2419} 

\ificcvfinal\fi
\begin{document}

\title{UM-Adapt: Unsupervised Multi-Task Adaptation Using Adversarial Cross-Task Distillation}

\author{Jogendra Nath Kundu \qquad Nishank Lakkakula \qquad R. Venkatesh Babu\\
Video Analytics Lab, Indian Institute of Science, Bangalore, India\\
{\tt\small jogendrak@iisc.ac.in, nishank974@gmail.com, venky@iisc.ac.in}
}

\maketitle
\thispagestyle{empty}

\begin{abstract}
Aiming towards human-level generalization, there is a need to explore adaptable representation learning methods with greater transferability. Most existing approaches independently address task-transferability and cross-domain adaptation, resulting in limited generalization. In this paper, we propose UM-Adapt - a unified framework to effectively perform unsupervised domain adaptation for spatially-structured prediction tasks, simultaneously maintaining a balanced performance across individual tasks in a multi-task setting. To realize this, we propose two novel regularization strategies; a) Contour-based content regularization (CCR) and b) exploitation of inter-task coherency using a cross-task distillation module. Furthermore, avoiding a conventional ad-hoc domain discriminator, we re-utilize the cross-task distillation loss as output of an energy function to adversarially minimize the input domain discrepancy. Through extensive experiments, we demonstrate superior generalizability of the learned representations simultaneously for multiple tasks under domain-shifts from synthetic to natural environments. UM-Adapt yields state-of-the-art transfer learning results on ImageNet classification and comparable performance on PASCAL VOC 2007 detection task, even with a smaller backbone-net. Moreover, the resulting semi-supervised framework outperforms the current fully-supervised multi-task learning state-of-the-art on both NYUD and Cityscapes dataset.
\end{abstract}
\section{Introduction}

Deep networks have proven to be highly successful in a wide range of computer vision problems. They not only excel in classification or recognition based tasks, but also deliver comparable performance improvements for complex spatially-structured prediction tasks~\cite{eigen2015predicting} like semantic segmentation, monocular depth estimation etc. However, generalizability of such models is one of the major concerns before deploying them in a target environment, since such models exhibit alarming dataset or domain bias~\cite{chen2017no,kundu2019unsupervised}.
To effectively address this, researchers have started focusing on unsupervised domain adaptation approaches~\cite{csurka2017domain}.
In a fully-unsupervised setting without target annotations, one of the effective approaches~\cite{ghifary2015domain,tzeng2014deep} is to minimize the domain discrepancy at a latent feature level so that the model extracts domain agnostic and task-specific representations. Although such approaches are very effective in classification or recognition based tasks~\cite{tzeng2017adversarial}, they
yield suboptimal performance for adaptation of fully-convolutional architectures, which is particularly essential for spatial prediction tasks~\cite{zhang2017curriculum}. One of the major issues encountered in such scenarios is attributed to the spatially-structured high-dimensional latent representation in contrast to vectorized form~\cite{kundu2018adadepth}. Moreover, preservation of spatial-regularity, avoiding mode-collapse~\cite{salimans2016improved} becomes a significant challenge while aiming to adapt in a fully unsupervised setting.

\begin{figure}[!tbp]
\centering    
	\includegraphics[width=1.0\linewidth]{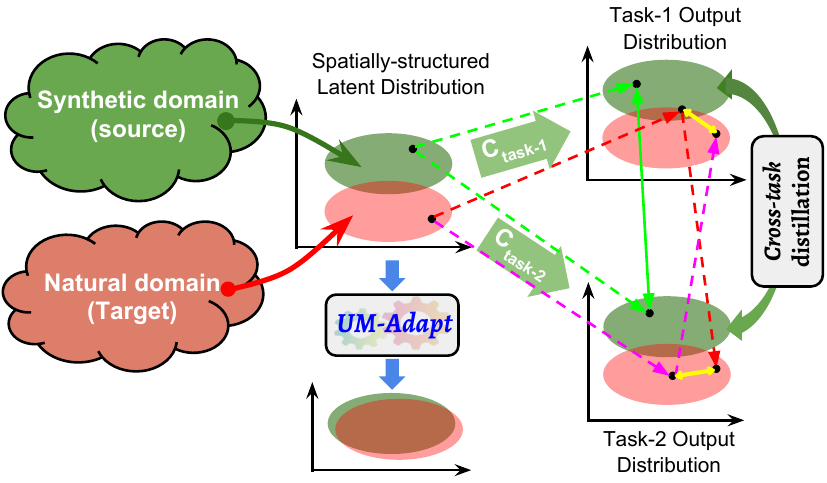}
	\caption{A schematic diagram to understand the implications of \textit{cross-task} distillation. The green arrows show consistency in \textit{cross-task} transfer for a source sample. Whereas, the red and purple arrows show a discrepancy (yellow arrows) in cross-task transfer for a target sample as a result of input domain-shift. \textit{UM-Adapt} aims to minimize this discrepancy as a proxy to achieve adaptation at a spatially-structured common latent representation.\vspace{-5mm}
	} 
	\label{fig_flow}    
\end{figure}




While we aim towards
human level performance, there is a need to explore scalable learning methods, which can yield generic image representation with improved transferability across both tasks and data domains. Muti-task learning~\cite{misra2016cross,kendall2017multi} is an emerging field of research in this direction, where the objective is to realize a task-agnostic visual representation by jointly training a model on several complementary tasks~\cite{hashimoto2016joint,long2015learning}. In general, such networks are difficult to train as they require explicit attention to balance performance across each individual task. 
Also, such approaches only address a single aspect of the final objective (i.e. generalization across tasks) ignoring the other important aspect of generalization across data domains. 

In this paper, we focus on multi-task adaptation of spatial prediction tasks, proposing efficient solutions to the specific difficulties discussed above. To effectively deliver optimal performance in both, generalization across tasks and input domain shift, we formulate a multitask adaptation framework called \textit{UM-Adapt}. 
To effectively preserve the spatial-regularity information during the unlabelled adversarial adaptation~\cite{tzeng2017adversarial} procedure, we propose two novel regularization techniques. Motivated by the fact that, the output representations share a common spatial-structure with respect to the input image, we first introduce a novel contour-based content regularization procedure. Additionally, we formalize a novel idea of exploiting cross-task coherency as an important cue to further regularize the multi-task adaptation process.

Consider a base-model trained on two different tasks, task-A and task-B. Can we use the supervision of task-A to learn the output representation of task-B and vice versa? Such an approach is feasible, particularly when the tasks in consideration share some common characteristics (i.e. consistency in spatial structure across task outputs). Following this reasoning, we introduce a \textit{cross-task Distillation} module (see Figure~\ref{fig_flow}). The module essentially constitutes of multiple encoder-decoder architectures called \textit{task-transfer} networks, which are trained to get back the representation of a certain task as output from a combination of rest of the tasks as the input. The overarching motivation behind such a framework is to effectively balance performance across all the tasks in consideration, thereby avoiding domination of the easier tasks during training. To intuitively understand the effectiveness of \textit{cross-task distillation}, let us consider a particular training state of the \textit{base-model}, where the performance on task-A is much better than the performance on task-B. Here the \textit{task-transfer} network (which is trained to output the task-B representation using the base-model task-A prediction as input) will yield improved performance on task-B, as a result of the dominant base task-A performance. This results in a clear discrepancy between the \textit{base-task-B} performance and the task-B performance obtained through the \textit{task-transfer} network. We aim to minimize this discrepancy which in turn acts as a regularization encouraging balanced learning across all the tasks.

In single-task domain adaptation approaches~\cite{tzeng2017adversarial,kundu2018adadepth}, it is common to employ an ad-hoc discriminator to minimize the domain discrepancy. However, in presence of \textit{cross-task distillation} module, this approach highly complicates the overall training pipeline. Therefore, avoiding such a direction, we propose to design a unified framework to effectively address both the diverse objectives i.e. a) to realize a balanced performance across all the tasks and b) to perform domain adaptation in an unsupervised setting. Taking inspirations from energy-based GAN~\cite{zhao2016energy}, we re-utilize the \textit{task-transfer} networks, treating the \textit{transfer-discrepancies} as output of energy-functions, to adversarially minimize the domain discrepancy in a fully-unsupervised setting.


Our contributions in this paper are as follows:
\begin{itemize}
    \item We propose a simplified, yet effective unsupervised multi-task adaptation framework, utilizing two novel regularization strategies; a) Contour-based content regularization (CCR) and b) exploitation of inter-task coherency using a \textit{cross-task} distillation module.

    \item Further, we adopt a novel direction by effectively utilizing \textit{cross-task} distillation loss as an energy-function to adversarially minimize the input domain discrepancy in a fully-unsupervised setting.

    \item\textit{UM-Adapt} yields \textit{state-of-the-art} transfer learning {results on ImageNet classification, and comparable performance} on PASCAL VOC 2007 detection task, even with a smaller backbone-net. The resulting semi-supervised framework outperforms the current fully-supervised multi-task learning \textit{state-of-the-art} on both NYUD and Cityscapes dataset.
\end{itemize}





\begin{figure*}\includegraphics[scale = 1.2]{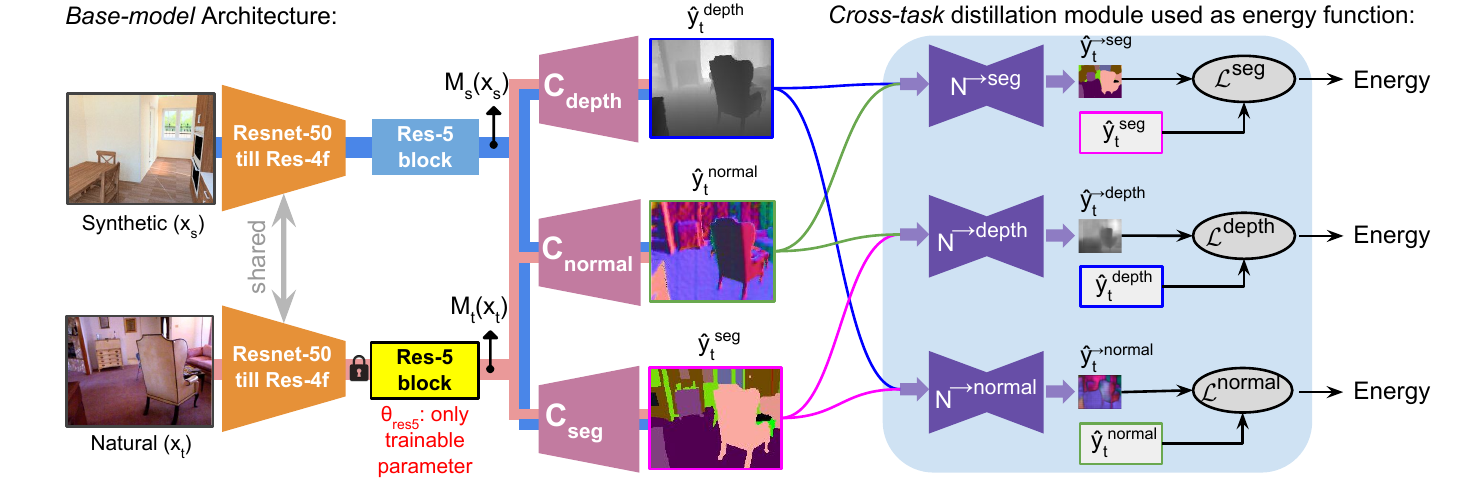}\centering\caption{An overview of the proposed \textit{UM-Adapt} architecture for multi-task adaptation. The blue and pink background wide-channel indicates data flow for synthetic and natural domain respectively. On the right we show an illustration of the proposed \textit{cross-task} distillation module, which is later utilized as an energy-function for adversarial adaptation (Section~\ref{section:3_3_2}).\vspace{-2mm}}
\label{fig_arch}
\end{figure*}

\section{Related work}

\noindent
\textbf{Domain adaptation.}
Recent adaptation approaches for deep networks focus on minimization of domain discrepancy by optimizing some distance function related to higher order statistical distributions~\cite{csurka2017domain}. Works following adversarial discriminative approaches ~\cite{tzeng2015simultaneous,tzeng2017adversarial,ganin2015unsupervised,ganin2016domain} utilize motivations from Generative Adversarial Networks~\cite{goodfellow2014generative} to bridge the domain-gap. 
Recent domain adaptation approaches, particularly targeting spatially-structured prediction tasks, can be broadly divided into two sub branches viz. a) pixel-space adaptation and b) feature space adaptation. In pixel-space adaptation ~\cite{atapour2018real,hoffman2017cycada,bousmalis} the objective is to train a image-translation network~\cite{zhu2017unpaired}, which can transform an image from the target domain to resemble like an image from the source domain. 
On the other hand, feature space adaptation approaches focus on minimization of various statistical distance metrics~\cite{gretton2009covariate,long2015learning,sun2016deep,hong2018conditional} at some latent feature-level, mostly in a fully-shared source and target domain parameter setting~\cite{ren2018cross}. However, unshared setups show improved adaptation performance as a result of learning dedicated filter parameters for both the domains in consideration~\cite{tzeng2017adversarial,kundu2018adadepth}. But, a fully unshared setup comes with other difficulties such as \textit{mode-collapse} due to inconsistent output in absence of paired supervision~\cite{isola2016image,zhu2017unpaired}. Therefore, an optimal strategy would be to adapt minimally possible parameters separately for the target domain in a partially-shared architecture setting~\cite{kundu2018adadepth}.

\vspace{1.1mm}
\noindent
\textbf{Multi-task learning.}
Multi-task learning~\cite{caruana1997multitask} has been applied in computer vision literature~\cite{misra2016cross} for quite a long time for a varied set of tasks in consideration~\cite{kruthiventi2016saliency,eigen2015predicting,eigen2014depth,han2017heterogeneous,ranjan2017hyperface}. To realize this, a trivial direction is to formulate a multi-task loss function, which weighs the relative contribution of each task, enabling equal importance to individual task performance. 
It is desirable to formulate techniques which adaptively modify the relative weighting of individual tasks as a function of the current learning state or iteration. Kendall~\etal~\cite{kendall2017multi} proposed a principled approach by utilizing a joint likelihood formulation to derive task weights based on the intrinsic uncertainty in individual tasks. Chen~\etal~\cite{Chen_2018} proposed a gradient normalization (GradNorm) algorithm that automatically balances training in deep multitask models by dynamically tuning gradient magnitudes. Another set of work, focuses on learning task-agnostic generalized visual representations utilizing the advances in multi-task learning techniques~\cite{yao2012describing,ren2018cross}.

\section{Approach}
\label{section:3}
Here we define the notations and problem setting for unsupervised multi-task adaptation. Consider source input image samples $x_s \in X_s$ with the corresponding outputs for different tasks being $y_s^{depth} \in Y_s^{depth}$, $y_s^{seg} \in Y_s^{seg}$ and $y_s^{normal} \in Y_s^{normal}$ for a set of three complementary tasks namely, monocular-depth, semantic-segmentation, and surface-normal respectively. We have full access to the source image and output pair as it is extracted from synthetic graphical environment. The objective of \textit{UM-Adapt} is 
to estimate the most reliable task-based predictions for an unknown target domain input,  $x_t \in X_t$. Considering natural images as samples from the target domain $P(X_t)$, to emulate an unsupervised setting, we restrict access to the corresponding task-specific outputs viz.  $y_t^{depth} \in Y_t^{depth}$, $y_t^{seg} \in Y_t^{seg}$ and $y_t^{normal} \in Y_t^{normal}$. Note that the objective can be readily extended to a semi-supervised setting, considering availability of output annotations for only few input samples from the target domain.

\subsection{UM-Adapt architecture}
\label{section:3_1}
As shown in Figure \ref{fig_arch}, the base multi-task adaptation architecture is motivated from the standard CNN encoder-decoder framework. The mapping function from the source domain, $X_s$ to a spatially-structured latent representation is denoted as $M_s(X_s)$. Following this, three different decoders with up-convolutional layers~\cite{laina2016deeper} are employed for the three tasks in consideration, (see Figure~\ref{fig_arch}) i.e., $\hat{Y}_s^{depth} = C_{depth}\circ M_s(X_s)$,  $\hat{Y}_s^{seg} = C_{seg}\circ M_s(X_s)$, and  $\hat{Y}_s^{normal} = C_{normal}\circ M_s(X_s)$. Initially the entire architecture is trained with full supervision on source domain data. To effectively balance performance across all the tasks, we introduce the \textit{cross-task Distillation} module as follows.

\subsubsection{Cross-task Distillation module}
\label{section:3_1_1}

This module aims to get back the representation of a certain task through a transfer function which takes a combined representation of all other tasks as input. Consider $T$ as the set of all tasks, i.e. $T=\{t_1,t_2,...,t_k\}$, where $k$ is the total number of tasks in consideration. For a particular task $t_i$, we denote $\hat{Y}_s^{t_i}$ as the prediction of the \textit{base-model} at the $t_i$ output-head. We denote the task specific loss function as $\mathcal{L}^{t_i}(.,.)$ in further sections of the paper. Here, \textit{task-transfer} network is represented as $N^{\rightarrow{t_i}}$, which takes a combined set $O_s^{t_i}=\{ Y_s^{t_j} \;\vert\; t_j\in T-\{t_i\}\}$ as the input representation and the corresponding output is denoted as ${Y}_s^{\rightarrow{t_i}} = N^{\rightarrow{t_i}}(O_s^{t_i})$. The parameters of $N^{\rightarrow{t_i}}$, $\theta^s_{N^{\rightarrow{t_i}}}$ are obtained by optimizing a task-transfer loss function, denoted by $\mathcal{L}^{t_i}({Y}_s^{\rightarrow{t_i}}, Y_s^{t_i})$ and are kept frozen in further stages of training. However, one can feed the predictions of \textit{base-model} through the \textit{task-transfer} network to realize another estimate of task $t_i$ represented as $\hat{Y}_s^{\rightarrow{t_i}} = N^{\rightarrow{t_i}}(\hat{O}_s^{t_i})$, where $\hat{O}_s^{t_i}=\{ \hat{Y}_s^{t_j}\; \vert \;t_j\in T - \{t_i\}\}$. Following this, we define the \textit{distillation-loss} for task $t_i$ as $\mathcal{L}^{t_i}(\hat{Y}_s^{\rightarrow{t_i}}, Y_s^{t_i})$. While optimizing parameters of the \textit{base-model}, this \textit{distillation-loss} is utilized as one of the important loss components to realize an effective balance across all the task objectives (see Algorithm~\ref{algo:1}). Here, we aim to minimize the discrepancy between the direct and indirect prediction (via other tasks) of individual tasks. The proposed learning algorithm does not allow any single task to dominate the training process, since the least performing task will exhibit higher discrepancy and hence will be given more importance in further training iterations.

Compared to the general knowledge-distillation framework~\cite{hinton2015distilling}, one can consider ${Y}_s^{\rightarrow{t_i}} = N^{\rightarrow{t_i}}(O_s^{t_i})$ to be analogous to the output of a teacher network and $\hat{Y}_s^{\rightarrow{t_i}} = N^{\rightarrow{t_i}}(\hat{O}_s^{t_i})$ as the output of a student network. Here the objective is to optimize the parameters of the \textit{base-model} by effectively employing the distillation loss which in turn enforces coherence among the individual task performances.

\begin{figure}[!t]
\includegraphics[width=0.9\linewidth]{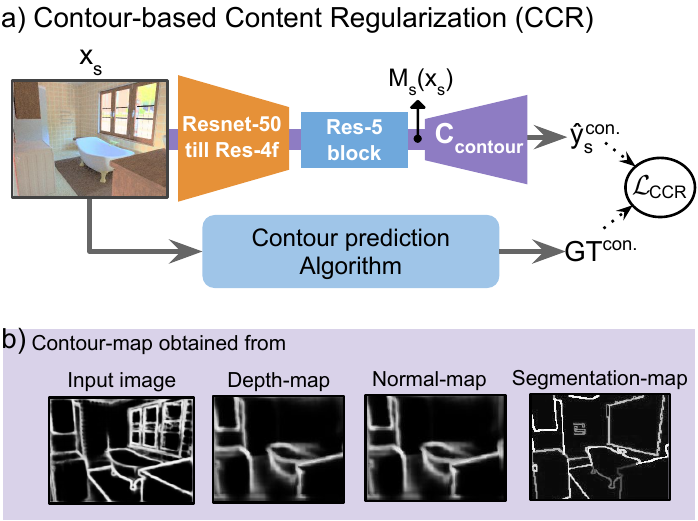}
\centering\caption{An overview of the (a) proposed CCR framework with (b) evidence of consistency in contour-map computed on the output-map of each task to that of the input RGB image.\vspace{-2mm}}
\label{fig_trn}
\end{figure}

\subsubsection{Architecture for target domain adaptation}
\label{section:3_1_2}

Following a partially-shared adaptation setup, a separate latent mapping network is introduced specifically for the target domain samples, i.e. $M_t(X_t)$ (see Figure~\ref{fig_arch}). Inline with AdaDepth~\cite{kundu2018adadepth}, we initialize $M_t$ using the pre-trained parameters of the source domain counterpart, $M_s$ (\textit{Resnet-50} encoder), in order to start with a good baseline initialization. Following this, only \textit{Res-5} block (i.e. $\theta_{res5}$) parameters are updated for the additional encoder branch, $M_t$. Note that the learning algorithm for unsupervised adaptation does not update other layers in the task-specific decoders and the initial shared $M_t$ layers till the \textit{Res-4f} block.

\begin{algorithm}[!t]
\label{algo:1}
\SetAlgoLined
        /*Initialization of parameters */\\
$\theta_{base}$: \textit{base-model} parameters $\{\theta_{M_s}, \theta_{T_{t_1}},...,\theta_{T_{t_k}}$\}\\

\For {$m$ iterations}{
 \For {task $t_i$; $i=1,2,...,k$}{
    $\mathcal{L}^{t_i}_{dist.} = \mathcal{L}^{t_i}(\hat{Y}_s^{t_i}, Y_s^{t_i}) + \alpha \mathcal{L}^{t_i}(\hat{Y}_s^{\rightarrow{t_i}}, Y_s^{t_i})$\\
    $\theta_{base}^* := \underset{\theta_{base}}{\textrm{argmin}}\hspace{1mm}\mathcal{L}^{t_i}_{dist.}$\\
 }
 }\vspace{1mm}
\caption{\textit{Base-model} training algorithm on fully-supervised source data with cross-task distillation loss. \vspace{-5mm}} 
\end{algorithm}

\subsection{Contour-based Content Regularization (CCR)}\label{section:3_2}
Spatial content inconsistency is a serious problem for unsupervised domain adaptation focused on pixel-wise dense-prediction tasks~\cite{hoffman2017cycada,kundu2018adadepth}. In order to address this,~\cite{kundu2018adadepth} proposes a Feature Consistency Framework (\textit{FCF}), where the authors employ a cyclic feature reconstruction setup to preserve the spatially-structured content such as semantic contours, at the \textit{Res-4f} activation map of the frozen \textit{Resnet-50} (till \textit{Res-4f}) encoder. However, the spatial size of the output activation of \textit{Res-4f} feature (i.e. 20$\times$16) is inefficient to capture relevant spatial regularities required to alleviate the contour alignment problem. 

To address the above issue, we propose a novel Contour-based Content Regularization (\textit{CCR}) method. As shown in Figure~\ref{fig_trn}a, we introduce a shallow (4 layer) contour decoder, $C_{contour}$ to reconstruct only the contour-map of the given input image, where the ground-truth is obtained using a standard contour prediction algorithm~\cite{xie15hed}. This content regularization loss (mean-squared loss) is denoted as  $\mathcal{L}_{CCR}$ in further sections of the paper. Assuming that majority of the image based contours align with the contours of task-specific output maps, we argue that the encoded feature (\textit{Res-5c} activation) must retain the contour information during the adversarial training for improved adaptation performance. This clearly makes \textit{CCR} superior over the existing approaches of enforcing image reconstruction based regularization~\cite{atapour2018real,murez2017image,sankaranarayanan2018learning}, by simplifying the additionally introduced decoder architecture devoid of the burden of generating irrelevant color-based appearance. $C_{contour}$ is trained on the fixed output transformation $M_s(X_s)$ and the corresponding ground-truth contour pair, i.e. $\mathcal{L}_{CCR}(\hat{Y}_s^{con.},GT^{con.})$. However, unlike \textit{FCF} regularization~\cite{kundu2018adadepth}, the parameter of $C_{contour}$ is not updated during the adversarial learning, as the expected output contour map is independent of the $M_s$ or $M_t$ transformation. As a result, $\mathcal{L}_{CCR}$ is treated as output of an energy-function, which is later minimized for $M_t(X_t)$ to bridge the discrepancy between the distributions $P(M_s(X_s))$ and $P(M_t(X_t))$ during adaptation, as shown in Algorithm~\ref{algoo:2}.


\subsection{Unsupervised Multi-task adaptation}\label{section:3_3}
In unsupervised adaptation, the overall objective is to minimize the discrepancy between the source and target input distributions. However, minimizing the discrepancy between $P(Y_s^{t_i})$ and $P(\hat{Y}_t^{t_i})$ can possibly overcome the differences between the ground-truth $Y_s^{t_i}$ and prediction $\hat{Y}_s^{t_i}$ better, when compared to matching $P(\hat{Y}_s^{t_i})$ with $P(\hat{Y}_t^{t_i})$ as proposed in some previous approaches~\cite{Tsai_adaptseg_2018}. Aiming towards optimal performance, \textit{UM-Adapt} focuses on matching target prediction with the actual ground-truth map distribution, and the proposed \textit{cross-task distillation} module provides a means to effectively realize such an objective.

\subsubsection{UM-Adapt baseline (\textit{UM-Adapt-B})}\label{section:3_3_1}
Existing literature~\cite{long2016unsupervised,kundu2018adadepth} shows efficacy of simultaneous adaptation at hierarchical feature levels, while minimizing domain discrepancy for multi-layer deep architectures. 
Motivated by this, we design a single discriminator which can match the joint distribution of latent representation and the final task-specific structured prediction maps with the corresponding true joint distribution. As shown in Figure~\ref{fig_arch}, the predicted joint distribution denoted by $P( M_t(X_t),\, \hat{Y}_t^{depth},\, \hat{Y}_t^{normal},\, \hat{Y}_t^{seg} )$, is matched with true distribution denoted by $P( M_s(X_s),\, Y_s^{depth},\, Y_s^{normal},\, Y_s^{seg} )$, following the usual adversarial discriminative strategy~\cite{kundu2018adadepth} (see Supplementary for more details).
We will denote this framework as \textit{UM-Adapt-B} in further sections of this paper.

\begin{algorithm}[!t]
\label{algoo:2}
\SetAlgoLined
        /*Initialization of parameters */\\
$\theta_{res5}$: \textit{Res5} parameters of $M_t$ initialized from $M_s$\\
$\theta_{N^{\rightarrow{t_i}}}$: parameters of fully trained $N^{\rightarrow{t_i}}$ (i.e. $\theta^s_{N^{\rightarrow{t_i}}}$)\\ \hspace{10mm} on ground-truth task output-maps, ${Y}_s^{t_i}$\\
\For {$n$ iterations}{
\For {$m$ steps}{
 \For {task $t_i$; $i=1,2,...,k$}{
    /* Update trainable parameters of $M_t$  by minimizing energy of target samples.*/\\
    $\mathcal{L}^{t_i}_G = \mathcal{L}^{t_i}(\hat{Y}_t^{\rightarrow{t_i}}, \hat{Y}_t^{t_i})$\\
    $\theta_{res5}^* := \underset{\theta_{res5}}{\textrm{argmin}}\hspace{1mm} \mathcal{L}^{t_i}_G + \lambda\mathcal{L}_{CCR}(\hat{Y}_t^{con.},GT^{con.})$\\
 }}
 \For {task $t_i$; $i=1,2,...,k$}{
    /* Update the energy function $N^{\rightarrow{t_i}}$ */\\
    $\mathcal{L}^{t_i}_D =  \mathcal{L}^{t_i}(\hat{Y}_s^{\rightarrow{t_i}}, {Y}_s^{t_i}) - \mathcal{L}^{t_i}(\hat{Y}_t^{\rightarrow{t_i}}, \hat{Y}_t^{t_i})$\\
    $\theta_{N^{\rightarrow{t_i}}}^* := \underset{\theta_{N^{\rightarrow{t_i}}}}{\textrm{argmin}}\hspace{1mm}\mathcal{L}^{t_i}_D$\\
 }
 }\vspace{1mm}
\caption{Training algorithm of \textit{UM-Adapt-(Adv.)} utilizing energy-based adversarial \textit{cross-task} distillation. In \textit{UM-Adapt-(noAdv.)} we do not update parameters of the t\textit{ask-transfer} network, i.e. $\theta^*_{N^{\rightarrow{t_i}}} = \theta_{N^{\rightarrow{t_i}}}$ throughout the adaptation procedure (see Section~\ref{section:3_3_2}).
\vspace{-4mm}}
\end{algorithm}

\subsubsection{Adversarial cross-task distillation}\label{section:3_3_2}
Aiming towards formalizing a unified framework to effectively address multi-task adaptation as a whole, we plan to treat the \textit{task-transfer} networks, $N^{\rightarrow{t_i}}$ as energy functions to adversarially minimize the domain discrepancy. Following the analogy of Energy-based GAN~\cite{zhao2016energy}, the \textit{task-transfer} networks are first trained to obtain low-energy for the ground-truth task-based source tuples (i.e. (${O}_s^{t_i}, {Y}_s^{t_i}$)) and high-energy for the similar tuples from the target predictions (i.e. ($\hat{O}_t^{t_i}, \hat{Y}_t^{t_i}$)). This is realized by minimizing $\mathcal{L}_D^{t_i}$ as defined in Algorithm~\ref{algoo:2}. Conversely, the trainable parameters of $M_t$ are updated to assign low energy to the predicted target prediction tuples, as enforced by $\mathcal{L}_G^{t_i}$ (see Algorithm~\ref{algoo:2}). Along with the previously introduced CCR regularization, the final update equation for $\theta_{res5}$ is represented as $\mathcal{L}^{t_i}_G + \lambda\mathcal{L}_{CCR}$. We use different optimizers for energy functions of each task, $t_i$. As a result, $\theta_{res5}$ is optimized to have a balanced performance across all the tasks even in a fully unsupervised setting. We denote this framework as \textit{UM-Adapt-(Adv.)} in further sections of this paper.


Note that the \textit{task-transfer} networks are trained only on ground-truth output-maps under sufficient regularization due to the compressed latent representation, as a result of the encoder-decoder setup.
This enables $N^{\rightarrow{t_i}}$ to learn a better approximation of the intended \textit{cross-task} energy manifold, even in absence of negative examples (target samples)~\cite{zhao2016energy}. This analogy is used in Algorithm~\ref{algo:1} to effectively treat the frozen \textit{task-transfer} network as an energy-function to realize a balanced performance across all the tasks on the fully-supervised source domain samples. Following this, we plan to formulate an ablation of \textit{UM-Adapt}, where we restrain the parameter update of $N^{\rightarrow{t_i}}$ in Algorithm~\ref{algoo:2}. We denote this framework as \textit{UM-Adapt-(noAdv.)} in further sections of this paper. This modification gracefully simplifies the unsupervised adaptation algorithm, as it finally retains only $\theta_{res5}$ as the minimal set of trainable parameters (with frozen parameter of $N^{\rightarrow{t_i}}$ as $\theta^s_{N^{\rightarrow{t_i}}}$).

\section{Experiments}\label{section:4}
To demonstrate effectiveness of the proposed framework, we evaluate on three different publicly available benchmark datasets, separately for indoor and outdoor scenes. Further, in this section, we discuss details of our adaptation setting and analysis of results on standard evaluation metrics for a fair comparison against prior art.

\subsection{Experimental Setting}\label{section:4_1}
We follow the encoder-decoder architecture exactly as proposed by Liana \etal~\cite{laina2016deeper}. The decoder architecture is replicated three times to form $C_{depth}$, $C_{normal}$ and $C_{seg}$ respectively. However, the number of feature maps and nonlinearity for the final task-based prediction layers is adopted according to the standard requirements. We use BerHu loss~\cite{laina2016deeper} as the loss function for the depth estimation task, i.e. $\mathcal{L}^{depth}(.,.)$. Following Eigen \etal~\cite{eigen2015predicting}, an inverse of element-wise dot product on unit normal vectors for each pixel location is consider as the loss function for surface-normal estimation, $\mathcal{L}^{normal}(.,.)$. Similarly for segmentation, i.e $\mathcal{L}^{seg}(.,.)$, classification based cross-entropy loss is implemented with a weighing scheme to balance gradients from different classes depending on their coverage.

We also consider a semi-supervised setting (\textit{UM-Adapt-S}), where the training starts from the initialization of the trained unsupervised version, \textit{UM-Adapt-(Adv.)}. For better generalization, alternate batches of labelled (optimize supervised loss, $\mathcal{L}_{dist.}^{t_i}$) and unlabelled (optimize unsupervised loss, $\mathcal{L}_{G}^{t_i}+\lambda\mathcal{L}_{CCR}$) target samples are used to update the network parameters (i.e. $\theta_{res5}$).

\begin{table}
 \centering
 \caption{Quantitative comparison of different ablations of \textit{UM-Adapt} framework with comparison against prior arts for depth estimation on NYUD-v2. The second column indicates amount of supervised target samples used during training.
 }
\setlength\tabcolsep{3pt}
\resizebox{1.01\linewidth}{!} {
\begin{tabular}{l|c|ccc|ccc}
\hline
\multirow{2}{*}{Method} & \multirow{2}{*}{\makecell{sup.}} & \multicolumn{3}{c|}{\tabincell{c}{Error$\downarrow$}} & \multicolumn{3}{c}{\tabincell{c}{Accuracy $\uparrow$ ($\gamma = 1.25$)}} \\ \cline{3-8}
                                 &     & rel & log10 & rms & $\delta < \gamma$ & $\delta<\gamma^2$ & $\delta < \gamma^3$ \\ \hline \hline
                                 
Saxena \etal~\cite{saxena2009make3d}     &    795            & 0.349 &  -     & 1.214  &0.447 & 0.745 & 0.897 \\

Liu \etal~\cite{liu2015deep}         &             795           & 0.230 &0.095& 0.824& 0.614& 0.883&  0.975 \\


Eigen \etal~\cite{eigen2014depth}  &        120K            & 0.215  &  -      & 0.907& 0.611 &  0.887  &  0.971\\
Roy~\etal~\cite{roy2016monocular}         &          795                & 0.187  &  0.078& 0.744& - &  -  &  -\\

Laina \etal~\cite{laina2016deeper}         &     96K       & 0.129  &0.056& {0.583}&0.801 & 0.950 & 0.986\\
 \hline  \multicolumn{8}{c}{Simultaneous multi-task learning}\\  \hline 
Multi-task baseline & 0 &0.27 &0.095	&0.862		&0.559	&0.852	&0.942 \\
UM-Adapt-B(FCF) & 0 &0.218	&0.091 &0.679		&0.67	&0.898	&0.974 \\
UM-Adapt-B(CCR) & 0 &0.192 &0.081	&0.754	&0.601	&0.877	&0.971 \\
UM-Adapt-(noAdv.)-1 & 0 &0.181	&0.077 &0.743 &0.623	&0.889	&0.978\\
UM-Adapt-(noAdv.) & 0 &0.178	&\textbf{0.063} &0.712 &{0.781}	&0.917	&0.984\\
UM-Adapt-(Adv.) & 0 &\textbf{0.175} &0.065	&\textbf{0.673}		&\textbf{0.783}	&\textbf{0.92}	&\textbf{0.984}\\ \hline
Wang \etal~\cite{wang2015towards}     &     795         & 0.220 & 0.094 &0.745& 0.605 & 0.890 & 0.970 \\
Eigen \etal~\cite{eigen2015predicting}   &    795        & 0.158  & -       & 0.641& 0.769 & 0.950 & 0.988 \\
Jafari~\etal~\cite{jafari2017analyzing} & 795 & 0.157 & 0.068 & 0.673 & 0.762 & 0.948 & 0.988 \\ 
UM-Adapt-S & 795 &\textbf{0.149}	&\textbf{0.067} &\textbf{0.637}		&\textbf{0.793}	&\textbf{0.938}	&\textbf{0.983} \\

\hline                        
\end{tabular}
}
\label{t_nyud_depth}
\vspace{-5pt}
\end{table}

\begin{table}
 \centering
 \caption{Quantitative comparison of different ablations of \textit{UM-Adapt} framework with comparison against prior arts for Surface-normal estimation on the standard test-set of NYUD-v2.
 }
\setlength\tabcolsep{3pt}
\resizebox{1.01\linewidth}{!} {
\begin{tabular}{l|c|cc|ccc}
\hline
\multirow{2}{*}{Method} & \multirow{2}{*}{\makecell{sup.}} & \multicolumn{2}{c|}{\tabincell{c}{Error $\downarrow$}} & \multicolumn{3}{c}{\tabincell{c}{Accuracy$\uparrow$}} \\ \cline{3-7}
                                 &     & mean & median & 11.25\degree  & 22.5\degree & 30\degree \\ \hline \hline
Eigen~\etal~\cite{eigen2015predicting} & 120k & 22.2	&15.3	&38.6	&64	&73.9 \\                 
PBRS~\cite{zhang2016physically}  & 795 & 21.74	&14.75	&39.37	&66.25	&76.06   \\     
SURGE~\cite{wang2015designing} & 795 &20.7 & 12.2 & 47.3 & 68.9 &76.6 \\
GeoNet~\cite{qi2018geonet} &30k &\textbf{19.0}	&\textbf{11.8}	&\textbf{48.4}	&71.5	&79.5 \\ 
 \hline  \multicolumn{7}{c}{Simultaneous multi-task learning}\\  \hline 
Multi Task Baseline&0	&25.8	&18.73	&29.65	&61.69	&69.83 \\
UM-Adapt-B(FCF)&0	&24.6	&16.49	&37.53	&65.73	&75.51 \\
UM-Adapt-B(CCR)&0	&23.8	&14.67	&42.08	&69.13	&77.28 \\
UM-Adapt-(noAdv.)-1&0	&22.3	&15.56	&43.17	&69.11	&78.36 \\
UM-Adapt-(noAdv.)&0	&22.2	&15.31	&\textbf{43.74}	&70.18	&78.83 \\
UM-Adapt-(Adv.)&0	&\textbf{22.2}	&\textbf{15.23}	&43.68	&\textbf{70.45}	&\textbf{78.95} \\
\hline
UM-Adapt-S &795	&21.2 &{13.98}	&44.66	&\textbf{72.11}	&\textbf{81.08} \\ \hline                        
\end{tabular}
}\label{t_nyud_normal}\vspace{-5pt}
\end{table}

\vspace{1.5mm}
\noindent
\textbf{Datasets.}
For representation learning on indoor scenes, we use the publicly available NYUD-v2~\cite{silberman2012indoor} dataset, which has been used extensively for supervised multi-task prediction of depth-estimation, sematic segmentation and surface-normal estimation. The processed version of the dataset consists of 1449 sample images with a standard split of 795 for training and 654 for testing. While adapting in semi-supervised setting, we use the corresponding ground-truth maps of all the 3 tasks (795 training images) for the supervised loss. The CNN takes an input of size 228$\times$304 with various augmentations of scale and flip following~\cite{eigen2014depth}, and outputs three task specific maps, each of size 128$\times$160. For the synthetic counterpart, we use 100,000 randomly sampled synthetic renders from PBRS~\cite{zhang2016physically} dataset along with the corresponding clean ground-truth maps (for all the three tasks) as the source domain samples.

To demonstrate generalizability of \textit{UM-Adapt}, we consider outdoor-scene dataset for two different tasks, sematic segmentation and depth estimation. For the synthetic source domain, we use the publicly available GTA5~\cite{richter2016playing} dataset consisting of 24966 images with the corresponding depth and segmentation ground-truths. However, for real outdoor scenes, the widely used KITTI dataset does not have semantic labels that are compatible with the synthetic counterpart. On the other hand, the natural image Cityscapes dataset~\cite{cordts2016cityscapes} does not contain ground-truth depth maps. Therefore to formulate a simultaneous multi-task learning problem and to perform a fair comparison against prior art, we consider the Eigen test-split on KITTI~\cite{eigen2015predicting} for comparison of depth-estimation result and the Cityscapes validation set to benchmark our outdoor segmentation results in a single \textit{UM-Adapt} framework. 
For the semi-supervised setting, we feed alternate KITTI and Cityscapes minibatches with the corresponding ground-truth maps for supervision. Here, input and output resolution for the network is considered to be 256$\times$512 and 128$\times$256 respectively.

\begin{table}[!t]
 \centering
 \caption{Quantitative comparison of different ablations of \textit{UM-Adapt} framework with comparison against prior arts for sematic segmentation on the standard test-set of NYUD-v2.
 }
\setlength\tabcolsep{3pt}
\resizebox{1.01\linewidth}{!} {
\begin{tabular}{l|c|ccc}
\hline Method & \makecell{Sup.}  & Mean IOU & Mean Accuracy & Pixel Accuracy \\                               
 \hline \hline
 PBRS~\cite{zhang2016physically}  & 795 & 0.332	& -	&-   \\  
Long~\etal~\cite{long2015fully} & 795 & 0.292	&0.422	&0.600 \\            
Lin~\etal~\cite{lin2016efficient} &795 &0.406	&0.536	&0.700 \\
Kong~\etal~\cite{kong2017recurrent} &795 &\textbf{0.445} & - & 0.721\\
RefineNet(Res50)~\cite{lin2017refinenet} & 795& 0.438 &- &-\\
 \hline  \multicolumn{5}{c}{Simultaneous multi-task learning}\\  \hline 
Multi Task Baseline&0	&0.022	&0.063	&0.067 \\
UM-Adapt-B(FCF)&0	&0.154	&0.295	&0.514 \\
UM-Adapt-B(CCR)&0	&0.163	&0.308	&0.557 \\
UM-Adapt-(noAdv.)-1&0	&0.189	&0.345	&0.603 \\
UM-Adapt-(noAdv.)&0	&0.214	&{0.364}	&0.608 \\
UM-Adapt-(Adv.)&0	&\textbf{0.221}	&\textit{0.366}	&\textbf{0.619} \\
\hline
Eigen~\etal~\cite{eigen2015predicting} &795 &0.341 &0.451 &0.656 \\
Arsalan~\etal~\cite{mousavian2016joint} &795 &0.392 &0.523 &0.686 \\
UM-Adapt-S &795	&\textbf{0.444}	&\textbf{0.536}	&\textbf{0.739} \\

\hline                        
\end{tabular}
}
\label{t_nyud_segment}
\vspace{-5pt}
\end{table}

\vspace{1.5mm}
\noindent
\textbf{Training details.}
We first train a set of task-transfer networks on the synthetic task label-maps separately for both indoor (PBRS) and outdoor (GTA5) scenes. For indoor dataset, we train only the following two \textit{task-transfer} networks; $N^{\rightarrow{seg}}({Y_s^{depth}, Y_s^{normal}})$ and $N^{\rightarrow{depth}}({Y_s^{seg}})$ considering the fact that surface-normal and depth estimation are more correlated, when compared to other pair of tasks.  Similarly for outdoor, we choose the only two {task-transfer} possible combinations $N^{\rightarrow{seg}}({Y_s^{depth}})$ and $N^{\rightarrow{depth}}({Y_s^{seg}})$. Following this, two separate \textit{base-model}s are trained with full-supervision on synthetic source domain using Algorithm~\ref{algo:1} ($\alpha=10$) with different optimizers (Adam~\cite{kingma2014adam}) for each individual task. After obtaining a frozen fully-trained source-domain network, the $C_{contour}$ network is trained as discussed in Section~\ref{section:3_2} and it remains frozen during its further usage as a regularizer. 

\subsection{Evaluation of the \textit{UM-Adapt} Framework}\label{section:4_2}
We have conducted a thorough ablation study to establish effectiveness of different components of the proposed \textit{UM-Adapt} framework. We report results on the standard benchmark metrics as followed in literature for each of the individual tasks, to have a fair comparison against state-of-the-art approaches. Considering the difficulties of simultaneous multi-task learning, we have clearly segregated prior art based on single-task or multi-task optimization approaches in all the tables in this section. 


\begin{table}
 \centering
 \caption{Quantitative comparison of ablations of \textit{UM-Adapt} framework with comparison against prior arts for depth-estimation on the Eigen test-split~\cite{eigen2014depth} of KITTI dataset.
 }
\setlength\tabcolsep{3pt}
\resizebox{1.01\linewidth}{!} {
\begin{tabular}{l|c|cccc}
\hline
\multirow{2}{*}{Method} & \multirow{2}{*}{\makecell{Target image\\ supervision}} &\multicolumn{4}{c}{\tabincell{c}{Error$\downarrow$}} 
\\ \cline{3-6} &     & rel & sq.rel & rms & rms(log10) 
                                 \\ \hline \hline
                                 
Eigen~\etal~\cite{eigen2014depth}& Full &0.203 &1.548 &6.307 &0.282  \\
Godard~\etal~\cite{godard2017unsupervised} & Binocular &0.148 &1.344 &5.927 &0.247 \\
zhou~\etal~\cite{zhou2017unsupervised} & Video &0.208 &1.768 &6.856 &0.283  \\
AdaDepth~\cite{kundu2018adadepth} &No &0.214 &1.932 &7.157 &0.295  \\
 \hline  \multicolumn{6}{c}{Simultaneous multi-task learning}\\  \hline 
Multi-task baseline & No &0.381	&2.08	&8.482	&0.41 \\

UM-Adapt-(noAdv.) & No &0.28	&1.99	&\textbf{7.791}	&0.346\\
UM-Adapt-(Adv.) & No &\textbf{0.27} &\textbf{1.98}	&7.823	&\textbf{0.336}\\ \hline
UM-Adapt-S & few-shot &\textbf{0.201} &\textbf{1.72} &\textbf{5.876} &\textbf{0.259} \\
\hline                        
\end{tabular}
}
\label{t_kitti_depth}
\vspace{-5pt}
\end{table}

\vspace{1.5mm}
\noindent
\textbf{Ablation study of \textit{UM-Adapt}.}\label{section:4_2_1}\hspace{2mm}
As a multi-task baseline, we report performance on the standard test-set of natural samples 
with direct inference on the frozen source-domain parameters without adaptation. With the exception of
results on sematic-segmentation, baseline performance for the other two regression tasks (i.e. depth estimation and surface-normal prediction) are strong enough to support the idea of achieving first-level generalization using multi-task learning. However, the prime focus of \textit{UM-Adapt} is to achieve the second-level of generalization through unsupervised domain adaptation. In this regard, to analyze effectiveness of the proposed \textit{CCR} regularization (Section~\ref{section:3_2}) against FCF~\cite{kundu2018adadepth}, we conduct an experiment on the \textit{UM-Adapt-B} framework defined in Section~\ref{section:3_3_1}. The reported benchmark numbers for unsupervised adaptation from PBRS to NYUD (see Table~\ref{t_nyud_depth},~\ref{t_nyud_normal} and~\ref{t_nyud_segment}) clearly indicates the superiority of \textit{CCR} for adaptation of structured prediction tasks. Following this inference, all later ablations (i.e. \textit{UM-Adapt-(noAdv.)}, \textit{UM-Adapt-(Adv.)} and \textit{UM-Adapt-S}) use only \textit{CCR} as  content-regularizer.

\begin{table}
 \centering
 \caption{Quantitative comparison of ablations of \textit{UM-Adapt} framework with comparison against prior arts for sematic segmentation on the validation set of Cityscapes dataset.
 }
\setlength\tabcolsep{17pt}
\resizebox{1.01\linewidth}{!} {
\begin{tabular}{l|c|c}
\hline Method & \makecell{Image supervision}  & Mean IOU \\   \hline \hline
FCN-Wild~\cite{hoffman2016fcns} & 0 & 0.271 \\
CDA~\cite{zhang2017curriculum} & 0 & 0.289 \\
DC~\cite{tzeng2015simultaneous} & 0 & 0.376 \\
Cycada~\cite{hoffman2017cycada} & 0 & 0.348 \\
AdaptSegNet~\cite{Tsai_adaptseg_2018} & 0 & 0.424 \\
 \hline  \multicolumn{3}{c}{Simultaneous multi-task learning}\\  \hline 
Multi Task Baseline &0	&0.224 \\
UM-Adapt-(noAdv.) &0	&0.408 \\
UM-Adapt-(Adv.) &0	&\textbf{0.420} \\
\hline
UM-Adapt-S &500 &\textbf{0.544} \\

\hline                        
\end{tabular}
}
\label{t_city_seg}
\vspace{-5pt}
\end{table}

Utilizing gradients from the frozen \textit{task-transfer} network yields a clear improvement over \textit{UM-Adapt-B} as shown in Tables~\ref{t_nyud_depth},~\ref{t_nyud_normal} and~\ref{t_nyud_segment} for all the three tasks in NYUD dataset. This highlights significance of the idea to effectively exploit the inter-task correlation information for adaptation of a multi-task learning framework. To quantify the importance of multiple \textit{task-transfer} network against employing a single such network, we designed another ablation setting denoted as \textit{UM-Adapt-(noAdv.)-1}, which utilizes only $N^{\rightarrow{seg}}(Y_s^{depth},Y_s^{normal})$ for adaptation to NYUD, as reported in Tables~\ref{t_nyud_depth},~\ref{t_nyud_normal} and~\ref{t_nyud_segment}.  Next, we report a comparison between the proposed energy-based cross-task distillation frameworks (Section~\ref{section:3_3_2}) i.e. a) \textit{UM-Adapt-(Adv.)} and b) \textit{UM-Adapt-(noAdv.)}. Though, \textit{UM-Adapt-(Adv.)} shows minimal improvement over the other counterpart,  training of \textit{UM-Adapt-(noAdv.)} is found to be significantly stable and faster as it does not include parameter update of the \textit{task-transfer} networks during the adaptation process.



\begin{figure*}\includegraphics[scale = 0.52]
{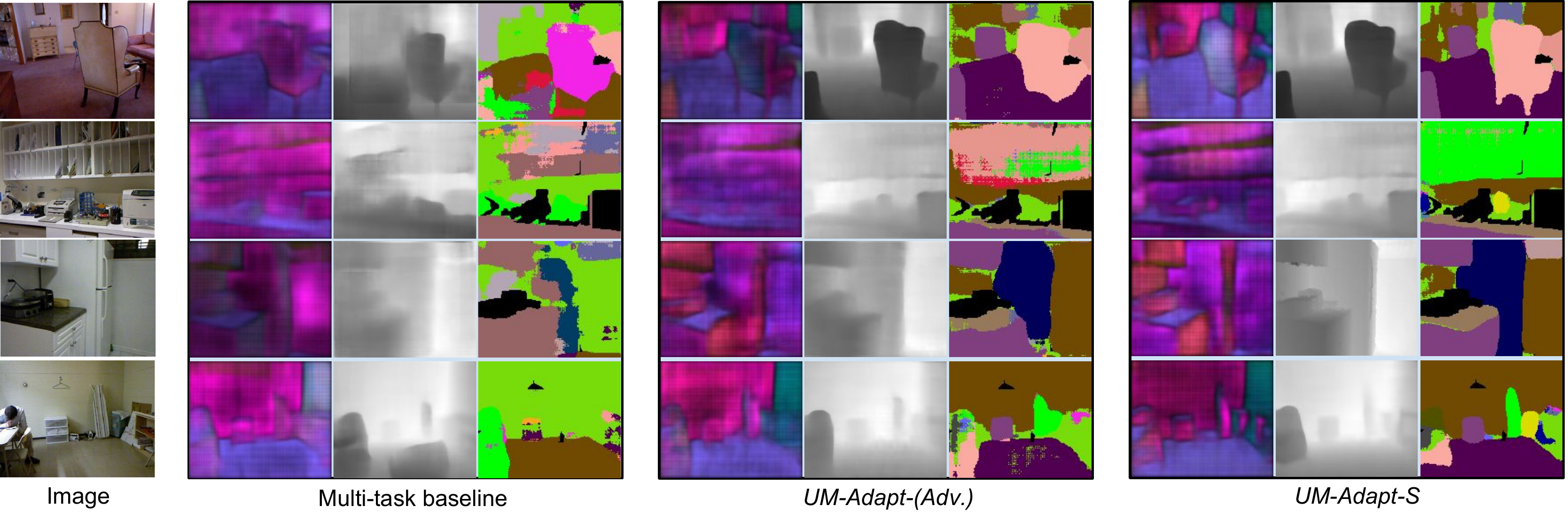}\centering\caption{Qualitative comparison of different ablations of \textit{UM-Adapt}, i.e. a) Multi-task baseline, b) \textit{UM-Adapt-(Adv.)}, and c) \textit{UM-Adapt-S}. 
}\vspace{-1mm}
\label{fig_nyud}\end{figure*}

\vspace{1.5mm}
\noindent
\textbf{Comparison against prior structured-prediction works.}\hspace{2mm}\label{section:4_2_2}
The final unsupervised multi-task adaptation result by the best variant of \textit{UM-Adapt} framework, i.e. \textit{UM-Adapt-(Adv.)} delivers comparable performance against previous fully-supervised approaches (see Table~\ref{t_nyud_depth} and~\ref{t_nyud_segment}). 
One must consider the explicit challenges faced by \textit{UM-Adapt} to simultaneously balance performance across multiple tasks in a unified architecture as compared to prior arts focusing on single-task at a time. This clearly demonstrates superiority of the proposed approach towards the final objective of realizing generalization across both tasks and data domains. The semi-supervised variant, \textit{UM-Adapt-S} is able to achieve state-of-the-art multi-task learning performance when compared against other fully supervised approaches as clearly highlighted in Tables~\ref{t_nyud_depth},~\ref{t_nyud_normal} and~\ref{t_nyud_segment}.

Note that the adaptation of depth estimation from KITTI and sematic-segmentation from Cityscapes in a single \textit{UM-Adapt} framework is a much harder task due to the input domain discrepancy (cross-city~\cite{chen2017no}) along with the challenges in simultaneous multi-task optimization setting. Even in such a drastic scenario \textit{UM-Adapt} is able to achieve reasonable performance in both depth estimation and sematic segmentation as compared to other unsupervised single task adaptation approaches reported in Tables~\ref{t_kitti_depth} and~\ref{t_city_seg}.

\vspace{1.5mm}
\noindent
\textbf{Comparison against prior multi-task learning works.}\label{section:4_2_3}\hspace{2mm}
Table~\ref{rt_1} and Table~\ref{rt_2} present a comparison of \textit{UM-Adapt} with recent multi-task learning approaches~\cite{kendall2017multi,Chen_2018} on NYUD test-set and CityScapes validation-set respectively. It clearly highlights ~\textit{state-of-the-art} performance achieved by \textit{UM-Adapt-S} as a result of the proposed \textit{cross-task} distillation framework.

\begin{table}[!h]
 \centering
 \caption{Test error on NYUDv2 with ResNet as the base-model
 \vspace{0.5mm}}
\setlength\tabcolsep{3pt}
\resizebox{1.01\linewidth}{!} {
\begin{tabular}{l|c|c|c|c}
\hline {Method} & {Sup.} &\makecell{Depth rms \\Err. (m)} & \makecell{Seg. Err.\\(100-IoU)} & \makecell{Normals Err.\\(1-$\vert cos\vert$)}\\ \hline
                                 
Kendall~\etal~\cite{kendall2017multi}& 30k &0.702 & - &0.182  \\
GradNorm~\cite{Chen_2018} & 30k &0.663 &67.5 &0.155 \\
\textit{UM-Adapt-S} & 795 &\textbf{0.637} &\textbf{55.6} &\textbf{0.139}\\
\hline                        
\end{tabular}
}
\label{rt_1}
\end{table}

\begin{table}[!h]
 \centering
  \vspace{-3mm}
 \caption{Validation mIOU on Cityscapes, where all the approaches are trained simultaneously for segmentation and depth estimation.
 }
\setlength\tabcolsep{3pt}
\resizebox{1.01\linewidth}{!} {
\begin{tabular}{l|c}
\hline {Method} &Semantic (mean IOU)\\ \hline
                                 
Kendall~\etal~\cite{kendall2017multi} (Uncert. Weights)& 51.52  \\
Liu~\etal~\cite{liu2018end} & 52.68 \\
\textit{UM-Adapt-S} &\textbf{54.4}\\
\hline                        
\end{tabular}
}
\label{rt_2}
\vspace{-2mm}
\end{table}

\vspace{1.5mm}
\noindent
\textbf{Transferability of the learned representation.}\label{section:4_2_4}\hspace{2mm}
One of the overarching goals of \textit{UM-Adapt} is to learn general-purpose visual representation, which can demonstrate improved transferability across both tasks and data-domains. 
To evaluate this, we perform experiments on large-scale representation learning benchmarks. Following evaluation protocol by Doersch~\etal~\cite{doersch2017multi}, we setup ~\textit{UM-Adapt} for transfer-learning on ImageNet~\cite{russakovsky2015imagenet} classification and PASCAL VOC 2007 Detection tasks. The base trunk till \textit{Res5} block is initialized from our ~\textit{UM-Adapt-(Adv.)} variant (the adaptation of PBRS to NYUD) for both classification and detection task. 
For ImageNet classification, we train randomly initialized fully connected layers after the output of \textit{Res5} block. Similarly, for detection we use \textit{Faster-RCNN}~\cite{ren2015faster} with 3 different output heads for object proposal, classification, and localization after the \textit{Res4} block. We finetune all the network weights separately for classification and detection~\cite{doersch2017multi}. 
The results in Table~\ref{rt_3} clearly highlight superior transfer learning performance of our learned representation even for novel unseen tasks with a smaller backbone-net. 

\begin{table}
\centering
 \caption{Transfer learning results on novel unseen tasks.
 }
\centering
\setlength\tabcolsep{3pt}
\resizebox{1.01\linewidth}{!} {
\begin{tabular}{l|c|c|c}
\hline {Method} &Backbone &\makecell{Classification\\ImageNet top5} & \makecell{Detection\\PASCAL 2007}\\ \hline
                                 
Motion Seg.~\cite{pathak2017learning}&ResNet-101 & 48.29 & 61.13 \\
Exemplar~\cite{doersch2017multi}&ResNet-101 & 53.08 & 60.94 \\
RP+Col+Ex+MS~\cite{doersch2017multi} &ResNet-101 & 69.30 & \textbf{70.53} \\
\textit{UM-Adapt-S} &\textbf{ResNet-50} & \textbf{69.51} & 70.02 \\
\hline                        
\end{tabular}
}
\label{rt_3}
\vspace{-5pt}
\end{table}

\section{Conclusion}\label{section:5}

The proposed \textit{UM-Adapt} framework addresses two important aspects of generalized feature learning by formulating the problem as a multi-task adaptation approach. While the multi-task training ensures learning of task-agnostic representation, the unsupervised domain adaptation method provides domain agnostic representation projected to a common spatially-structured latent representation. The idea of exploiting cross-task coherence as an important cue for preservation of spatial regularity can be utilized in many other scenarios involving fully-convolutional architectures. Exploitation of auxiliary task setting is another direction that remains to be explored in this context. 

\vspace{2mm}
\noindent
\textbf{Acknowledgements.} This work was supported by a CSIR Fellowship (Jogendra) and a grant from RBCCPS, IISc. We also thank Google India for the
travel grant.

{\small
\bibliographystyle{ieee_fullname}
\bibliography{egbib}
}




\section*{Supplementary material}

Table 1: Network architecture of the discriminator used in *UM-Adapt-B* baseline variant. Conv* denotes standard convolutional layers followed by a batch-normalization with leaky-ReLU non-linearity. Here, || denotes channel-wise concatenation.

| Layer | input | Type | Filter | Stride | Output Size |
|---|---|---|---|---|---|
| $C1_d$ | depth-map | Conv | $3 \times 3$, 64 | 1 | $128 \times 160 \times 64$ |
| $C2_d$ | $C1_d$ | Conv* | $3 \times 3$, 128 | 2 | $64 \times 80 \times 128$ |
| $C1_n$ | normal-map | Conv | $3 \times 3$, 64 | 1 | $128 \times 160 \times 64$ |
| $C2_n$ | $C1_n$ | Conv* | $3 \times 3$, 128 | 2 | $64 \times 80 \times 128$ |
| $C3_{n+d}$ | $C2_n||C2_d$ | Conv | $3 \times 3$, 256 | 2 | $32 \times 40 \times 256$ |
| $C1_m$ | $M_s(x_s)$ or $M_t(x_t)$ | Deconv | $1 \times 1$, 512 | 1 | $8 \times 10 \times 512$ |
| $C2_m$ | $C1_m$ | Deconv* | $3 \times 3$, 512 | 1 | $16 \times 20 \times 512$ |
| $C3_m$ | $C1_m$ | Deconv* | $3 \times 3$, 256 | 2 | $32 \times 40 \times 256$ |
| $C1_s$ | seg-map | Conv | $3 \times 3$, 64 | 1 | $128 \times 160 \times 64$ |
| $C2_s$ | $C1_s$ | Conv* | $3 \times 3$, 128 | 2 | $64 \times 80 \times 128$ |
| $C3_s$ | $C2_s$ | Conv* | $3 \times 3$, 256 | 2 | $32 \times 40 \times 256$ |
| $C1_c$ | $C3_s||C3_{n+d}||C3_m$ | Conv | $1 \times 1$, 512 | 1 | $32 \times 40 \times 512$ |
| $C2_c$ | $C1_c$ | Conv* | $3 \times 3$, 1024 | 2 | $16 \times 20 \times 1024$ |
| $C3_c$ | $C2_c$ | Conv* | $3 \times 3$, 512 | 2 | $8 \times 10 \times 512$ |
| $C4_c$ | $C3_c$ | Conv | $1 \times 1$, 1 | 1 | $8 \times 10 \times 1$ |

concatenation with the latent feature map followed by some deconvolutional layer to form an hour-glass like patch discriminator [3]. The discriminator architecture for *UM-Adapt-B* is presented in Table 1. Here the discriminator is trained following the improved techniques of Wasserstein GAN as proposed by Gulrajani *et al.* [2].

## Architecture of Task-transfer networks

As mentioned in the main paper, for NUYD we take a combined input of depth-map and normal-map for the baseline variant *UM-Adapt-(noAdv.)-1* (See Table 2). Whereas in *UM-Adapt-(noAdv.)* and *UM-Adapt-(Adv.)*, we use an additional task-transfer network which predicts depth-map output with segmentation map as the input representation. Note that, all the task-transfer networks are independently trained on synthetic depth, normal and segmentation tuples and kept frozen throughout the adaptation process for both *UM-Adapt-(noAdv.)-1* and *UM-Adapt-(noAdv.)*. Whereas for *UM-Adapt-(Adv.)*, we also update the parameters of the task-transfer networks as shown in Algorithm 2 (main paper). A similar task-transfer architecture is employed for GTA5 to KITTI+Cityscapes adaptation, where the input resolution is taken as 128×256.

Table 2: Architectural details of Task-transfer networks. Conv* denotes standard convolutional layers followed by a batch-normalization with leaky-ReLU non-linearity. Here, || denotes channel-wise concatenation.

| Layer | input | Type | Filter | Stride | Output Size |
|---|---|---|---|---|---|
| $C1_d$ | depth-map | Conv | $3 \times 3, 64$ | 1 | $128 \times 160 \times 64$ |
| $C2_d$ | $C1_d$ | Conv* | $3 \times 3, 128$ | 2 | $64 \times 80 \times 128$ |
| $C1_n$ | normal-map | Conv | $3 \times 3, 64$ | 1 | $128 \times 160 \times 64$ |
| $C2_n$ | $C1_n$ | Conv* | $3 \times 3, 128$ | 2 | $64 \times 80 \times 128$ |
| $C3_m$ | $C2_n || C2_d$ | Conv | $3 \times 3, 256$ | 2 | $32 \times 40 \times 256$ |
| $C3_m$ | $C2_m$ | Conv* | $3 \times 3, 512$ | 2 | $16 \times 20 \times 512$ |
| $C4_m$ | $C3_m$ | Conv* | $3 \times 3, 512$ | 1 | $16 \times 20 \times 512$ |
| $EN_{depth}(y^{depth})$ | $C4_m$ | Conv | $3 \times 3, 1024$ | 2 | $8 \times 10 \times 1024$ |
| $C5_m$ | $EN_{depth}(y^{depth})$ | Deconv* | $3 \times 3, 512$ | 2 | $16 \times 20 \times 512$ |
| $C6_m$ | $C5_m$ | Deconv* | $3 \times 3, 512$ | 2 | $32 \times 40 \times 512$ |
| $C7_m$ | $C6_m$ | Deconv* | $3 \times 3, 256$ | 2 | $64 \times 80 \times 256$ |
| $C8_m$ | $C7_m$ | Deconv* | $3 \times 3, 128$ | 2 | $128 \times 160 \times 128$ |
| $C9_{seg}$ | $C8_m$ | Conv(softmax) | $1 \times 1, 40$ | 1 | $128 \times 160 \times 40$ |

## Qualitative Results on KITTI and Cityscapes

We provide qualitative results on KITTI and Cityscapes dataset for both depth estimation and sematic segmentation tasks (See Figure 2). Note that in the semi-supervised setting we have only depth-map ground-truth for KITTI and only segmentation map ground-truth for Cityscapes dataset. We treat the mixed image samples from KITTI and Cityscapes together as the target domain input ignoring the cross-city [1] input discrepancy.

## Other Implementation Details

In contrast to depth-map or normal-map, the segmentation map ground-truth has a specific structure as it is represented as a one-hot vector for each pixel-location. Discriminating such one-hot real segmentation maps (synthetic data) from the output of *UM-Adapt* segmentation after softmax non-linearity becomes an easier task for the *task-transfer* networks or the discriminator, taking these representations as input at different iterations. To alleviate this, we create a random *pseudo-softmax* representation of the ground-truth one-hot maps by keeping higher probability for the true-class (sampled from a truncated Normal pdf with mean 1 and standard-deviation 0.075) in the randomly generated vector satisfying sum one and values in the range of zero to one criteria. We provide 50% of direct one-hot and 50% of such *pseudo-softmax* (at each batch iteration) for the true segmentation map distribution. Besides this, to address the problem of class-balancing, we use a randomly generated gradient mask layer (50% of the batch without masking) in-between the UM-Adapt prediction and the corresponding Task-transfer or the discriminator network.

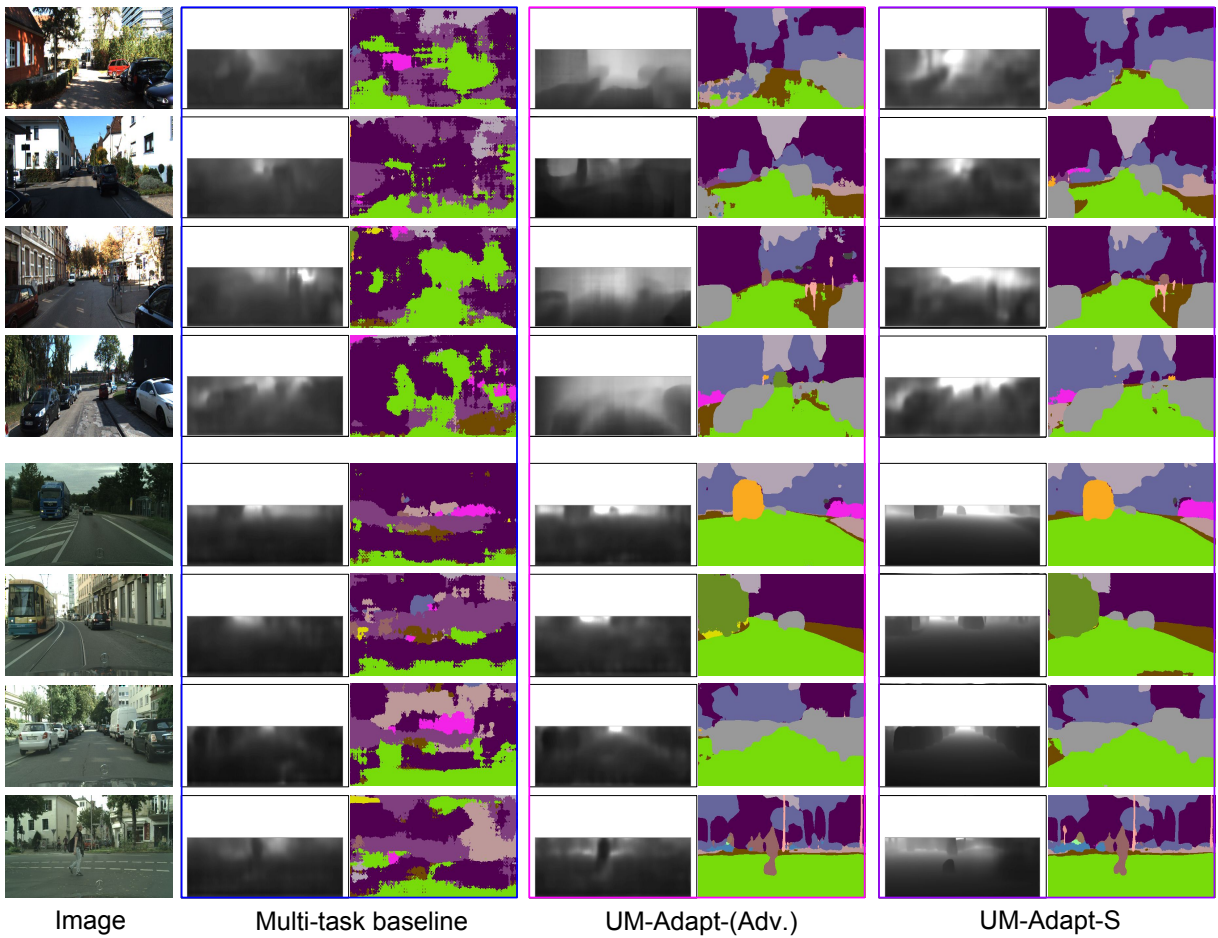

Figure 2: Qualitative comparison of different ablations of *UM-Adapt*. First 4 rows contains images from KITTI test set (Eigen-spilt), whereas the last 4 rows contains images from Cityscapes validation-set. We adapt a single multi-task model on combined training set from both the datasets.



{\small
\bibliographystyle{ieee_fullname}
\bibliography{egbib}
}

\end{document}